\documentclass[11pt, a4paper, logo, copyright, nonumbering]{deepseek}
\usepackage[authoryear, sort&compress, round]{natbib}
\usepackage{dblfloatfix}
\usepackage{ulem}
\usepackage{caption}
\usepackage{amsthm}

\usepackage{subcaption}
\usepackage{graphicx}
\usepackage{makecell}
\usepackage[utf8]{inputenc}
\usepackage{dramatist}
\usepackage{xspace}
\usepackage{pifont}
\usepackage{multirow}
\usepackage{tcolorbox}
\usepackage{tabularx}
\usepackage{xltabular}
\usepackage{longtable}
\usepackage{hyperref}
\AtBeginDocument{

}
\hypersetup{
    colorlinks=true,
    linkcolor=blue,
    citecolor=blue,
    urlcolor=blue,
    filecolor=blue
}
\usepackage{xcolor}
\usepackage{array}
\usepackage{amsfonts}
\usepackage{amsmath}
\usepackage{amssymb}
\usepackage{lineno}
\usepackage{bm}
\usepackage{booktabs}
\usepackage{ragged2e}
\usepackage{algorithm}
\usepackage{algorithmic}

\usepackage[bottom]{footmisc}

\usepackage{CJKutf8}
\usepackage{setspace}

\usepackage{dsfont}

\makeatletter
\def\@BTrule[#1]{%
  \ifx\longtable\undefined
    \let\@BTswitch\@BTnormal
  \else\ifx\hline\LT@hline
    \nobreak
    \let\@BTswitch\@BLTrule
  \else
     \let\@BTswitch\@BTnormal
  \fi\fi
  \global\@thisrulewidth=#1\relax
  \ifnum\@thisruleclass=\tw@\vskip\@aboverulesep\else
  \ifnum\@lastruleclass=\z@\vskip\@aboverulesep\else
  \ifnum\@lastruleclass=\@ne\vskip\doublerulesep\fi\fi\fi
  \@BTswitch}
\makeatother

\addto\extrasenglish{
}

 {\begin{list}{}%
         {\setlength{\leftmargin}{#1}}%
         \item[]%
 }
 {\end{list}}
 
\reportnumber{001} 

\renewcommand{\today}{}

\title{
\centering DSpark: Confidence-Scheduled Speculative Decoding with Semi-Autoregressive Generation
}

\author[]{
Xin Cheng$^{1,2,*}$, Xingkai Yu$^{2,*}$, Chenze Shao$^{2,*}$, Jiashi Li$^{2,*}$, Yunfan Xiong$^{2,*}$

\vspace{-0.1in}
Yi Qian$^{2}$, Jiaqi Zhu$^{2}$, Shirong Ma$^{2}$, Xiaokang Zhang$^{2}$, Jiasheng Ye$^{2}$, Qinyu Chen$^{2}$,

\vspace{-0.1in}
Chengqi Deng$^{2}$, Jiping Yu$^{2}$, Damai Dai$^{2}$,  Zhengyan Zhang$^{2}$, Yixuan Wei$^{2}$, Yixuan Tan$^{2}$,

\vspace{-0.1in}
Wenkai Yang$^{2}$, Runxin Xu$^{2}$, Yu Wu$^{2}$, Zhean Xu$^{2}$, Xuanyu Wang$^{2}$, Muyang Chen$^{2}$,

\vspace{-0.1in}
Rui Tian$^{2}$,  Xiao Bi$^{2}$, Zhewen Hao$^{2}$, Shaoyuan Chen$^{2}$, Huanqi Cao$^{2}$, Wentao Zhang$^{2}$,

\vspace{-0.1in}
Anyi Xu$^{2}$, Huishuai Zhang$^{1}$, Dongyan Zhao$^{1}$, Wenfeng Liang$^{2}$ \\
\vspace{-0.1in}
\small
$^1$Peking University  \quad $^2$DeepSeek-AI \\
\small
\{chengxin, xingkai, shaochenze, js.li, yunfanxiong\}@deepseek.com

\vspace{-0.2in}
}

\newcommand{\RR}{\mathbb{R}}

\newcommand{\Ll}{\mathcal{L}}

\renewcommand{\phi}{\varphi}

\renewcommand{\leq}{\leqslant}

\renewcommand{\epsilon}{\varepsilon}
\renewcommand{\imath}{\mathrm{i}}

\newlength{\restsubwidth}
\newlength{\restsubheight}
\newlength{\restsubmoreheight}
\newcommand{\rest}[2]{%
        \settowidth{\restsubwidth}{\ensuremath{#2}}
        \settoheight{\restsubheight}{\ensuremath{{}_{#2}}}
        \ensuremath{{#1\hskip 0.5pt}_{\vrule\kern2pt\parbox[b][%
        4pt][b]{\the\restsubwidth}{%
                        \ensuremath{{}_{#2}}}}}
        }

\begin{abstract}
Speculative decoding accelerates Large Language Model (LLM) inference by decoupling draft generation from target verification. While recent parallel drafters efficiently propose long token sequences in a single forward pass, they suffer from rapid acceptance decay due to a lack of inter-token dependencies. Furthermore, indiscriminately verifying these extended blocks wastes critical batch capacity on tokens with high rejection risks, severely degrading throughput in high-concurrency serving systems. We introduce DSpark, a speculative decoding framework that unifies high-throughput parallel generation with adaptive, load-aware verification. To maintain draft quality, DSpark utilizes a semi-autoregressive architecture—coupling a parallel backbone with a lightweight sequential module—to introduce intra-block dependency modeling and mitigate suffix decay. To optimize system efficiency, DSpark employs confidence-scheduled verification, dynamically tailoring the verification length for each request based on estimated prefix survival probabilities and engine-specific throughput profiles. On offline benchmarks across diverse domains, DSpark substantially improves the accepted length over state-of-the-art autoregressive and parallel drafters. When deployed within the DeepSeek-V4 serving system under live user traffic, DSpark successfully mitigates verification waste. 
Compared to the established production baseline~(MTP-1), DSpark accelerates per-user generation speeds by 60\%--85\% at matched throughput levels. 
More importantly, by preventing severe throughput degradation under strict interactivity constraints, it enables performance tiers that were previously unattainable, shifting the Pareto frontier of our serving system.
To facilitate community progress, we open-source the \href{https://huggingface.co/deepseek-ai/DeepSeek-V4-Pro-DSpark/tree/main}{DSpark checkpoints} alongside \href{https://github.com/deepseek-ai/DeepSpec}{DeepSpec}, an algorithm-driven training repository for speculative decoding.
\end{abstract}

\begin{document}
\begin{CJK*}{UTF8}{gbsn}

\maketitle

\renewcommand{\thefootnote}{*}
\footnotetext{Equal contribution.}
\renewcommand{\thefootnote}{\arabic{footnote}}


\section{Introduction}
\label{sec:intro}

Large Language Models (LLMs) generate text autoregressively: each new token requires a full forward pass conditioned on all preceding tokens, making inference latency proportional to the output length.
The resulting low GPU utilization and high user-perceived waiting time constitute a primary bottleneck in production LLM serving, particularly for latency-sensitive scenarios such as real-time conversational assistants and multi-turn agentic workflows.
Speculative decoding~\citep{pmlr-v202-leviathan23a,chen2023accelerating} offers a principled solution: a lightweight \textit{draft model} proposes a block of candidate tokens, and the full-size \textit{target model} verifies the entire block in a single forward pass via rejection sampling, accepting the longest prefix consistent with the target distribution and appending one bonus token.
Because verification is parallel and the acceptance rule preserves the target distribution exactly, speculative decoding accelerates generation without any quality loss.

The design of the draft model governs the trade-off between drafting latency and acceptance rate. Early drafters are autoregressive~\citep{li-etal-2024-eagle,cheng2024recurrentdrafterfastspeculative}, conditioning each position on previously sampled tokens. However, their drafting latency grows linearly with the block size, forcing these methods to use short blocks and shallow architectures. To break this sequential bottleneck, parallel drafters~\citep{pmlr-v235-cai24b,liu2026dartdiffusioninspiredspeculativedecoding,chen2026dflash} have emerged as a compelling alternative: all draft positions are produced in a single forward pass, making drafting latency nearly independent of block size. This structural advantage theoretically allows parallel drafters to efficiently generate substantially longer draft blocks.

However, fully unlocking the potential of large parallel draft blocks introduces two critical bottlenecks—one in generation quality, and the other in system efficiency.
First, because parallel drafters predict each position independently, they cannot model inter-token dependencies within a block. This independence leads to multi-modal collisions and rapid acceptance decay at later positions~\citep{gu2018nonautoregressive,pmlr-v162-huang22m}. 
Second, determining the optimal verification length remains a challenge. While parallel generation easily produces long draft blocks, indiscriminately verifying all proposed tokens degrades system throughput, particularly under high-concurrency workloads~\citep{liu2024turbospec,hu2026echo}. The ideal verification length varies along two axes. 
On the data side, structured requests like code naturally sustain higher acceptance rates than open-ended chat~\citep{xia-etal-2024-unlocking,abramovich2026speed}. 
On the system side, verifying extra tokens is nearly free under light loads. Under heavy loads, however, verifying tokens with a high rejection risk occupies critical batch capacity that could otherwise serve other active requests~\citep{liu2024optimizing, wu-etal-2025-tetris}.

To address these bottlenecks, we introduce \textbf{DSpark}, a speculative decoding framework that unifies high-throughput parallel generation with adaptive, load-aware verification. At its core, DSpark is designed to resolve the inherent trade-offs in draft generation and verification through two complementary mechanisms. 
\begin{itemize}
    \setlength{\itemsep}{10pt}
    \item First, to overcome the lack of inter-token dependencies, DSpark adopts a semi-autoregressive architecture. It keeps the computationally expensive draft backbone fully parallel, appending only a lightweight serial output head to inject local transition information. This design preserves the drafting speed of parallel models while significantly mitigating suffix decay.
    \item Second, to resolve the system-level bottleneck, DSpark employs confidence-scheduled verification. By coupling a confidence head—which estimates per-position prefix survival probabilities—with a hardware-aware scheduler, DSpark dynamically tailors the verification length for each request. This scheduler leverages real-time engine throughput profiles to route target verification budget only toward tokens with the highest expected return.
\end{itemize}

We extensively evaluate DSpark across both controlled offline benchmarks and production-scale online deployments.
On controlled offline benchmarks—spanning mathematical reasoning, code generation, and daily chat—DSpark consistently outperforms strong baselines. Specifically, across the Qwen3-4B, 8B, and 14B target models~\citep{yang2025qwen3}, it improves the macro-average accepted length over the autoregressive Eagle3~\citep{li2026eagle} by 30.9\%, 26.7\%, and 30.0\%, and over the parallel DFlash~\citep{chen2026dflash} by 16.3\%, 18.4\%, and 18.3\%, respectively. Beyond top-line metrics, our fine-grained position-wise analysis reveals the distinct generation characteristics of different drafters, empirically demonstrating how DSpark successfully combines the high initial-token capacity of parallel models with the suffix coherence of autoregressive models.

Beyond offline evaluation, we deployed DSpark within the DeepSeek-V4~\citep{deepseekai2026deepseekv4highlyefficientmilliontoken} serving system to assess its performance under live user traffic. Compared to the prior MTP-1 production baseline~\citep{liu2024deepseek}, DSpark significantly broadens the system's operational envelope. Specifically, it consistently accelerates per-user generation speeds by 60\%--85\% (V4-Flash) and 57\%--78\% (V4-Pro) at matched aggregate throughput capacities. 
Furthermore, under strict Service Level Agreements (SLAs) where the baseline's capacity deteriorates severely—such as 120 TPS for Flash and 50 TPS for Pro—DSpark mitigates verification overhead to maintain robust throughput. By overcoming this performance cliff, DSpark unlocks strict interactivity tiers that were previously unattainable, effectively shifting the Pareto frontier of LLM serving.

To foster collective advancement within the open-source community, we are making our artifacts publicly available. Specifically, we release the trained \href{https://huggingface.co/deepseek-ai/DeepSeek-V4-Pro-DSpark/tree/main}{DSpark checkpoints} for both the DeepSeek-V4-Flash~(preview) and DeepSeek-V4-Pro~(preview) models. Furthermore, we open-source \href{https://github.com/deepseek-ai/DeepSpec}{DeepSpec}, an algorithm-driven training repository, including Eagle3, DFlash and DSpark. 
These artifacts are intended to support further research on efficient LLM serving.
\section{Background}
\label{sec:background}

\subsection{Speculative Decoding}
\label{sec:spec_decoding}

Autoregressive language models generate one token per forward pass, making inference latency proportional to output length.
Speculative decoding~\citep{ge2022lossless,pmlr-v202-leviathan23a,chen2023accelerating} accelerates the inference of a target model $M_t$ using a lightweight draft model $M_d$.
At each decoding cycle, the draft model proposes $\gamma$ candidate tokens $x_1, \ldots, x_{\gamma}$. The target model verifies all candidates in a single forward pass, accepting the longest prefix consistent with its own distribution.

Concretely, at each draft position $k$, the target model computes its own distribution $p_k^t$ and compares it against the draft distribution $p_k^d$.
The token $x_k$ is accepted with probability $\min(1,\, p_k^t(x_k) / p_k^d(x_k))$.
Verification proceeds left to right: the first rejection at position $k$ discards all subsequent tokens $x_{k+1}, \ldots, x_\gamma$, regardless of their quality.

Let $\tau$ denote the number of accepted tokens per cycle, and let $T_{\text{draft}}$ and $T_{\text{verify}}$ be the wall-clock times of the drafting and verification passes, respectively.
The average latency per generated token is:
\begin{equation}
L = \frac{T_{\text{draft}} + T_{\text{verify}}}{\tau}.
\label{eq:latency}
\end{equation}
Improving speedup therefore reduces to three levers: lowering $T_{\text{draft}}$ (draft faster), raising $\tau$ (draft better), or reducing the effective $T_{\text{verify}}$ (verify smarter).

\subsection{Drafter Architectures}
\label{sec:drafter_arch}

The design of the draft model determines how $T_{\text{draft}}$ and $\tau$ trade off.
Existing approaches fall into two categories.

\paragraph{Autoregressive drafters.}
Autoregressive drafters generate draft tokens sequentially, conditioning each position on previously sampled tokens~\citep{liu2024deepseek,pmlr-v235-li24bt,li-etal-2024-eagle,li2026eagle,zhang2025learningharmonizedrepresentationsspeculative}.
This explicit dependency gives strong modeling capacity, but the drafting cost grows linearly with block size: $T_{\text{draft}} \propto \gamma$, which forces autoregressive drafters to use small $\gamma$ and shallow architectures to keep $T_{\text{draft}}$ low.
To compensate for the short block, tree-based verification~\citep{miao2024specinfer} expands candidates into a tree and verifies multiple paths via tree attention, but the large number of verification tokens reduces overall serving throughput.

\paragraph{Parallel drafters.}
Parallel drafters produce all $\gamma$ draft tokens in a single forward pass, making $T_{\text{draft}}$ nearly independent of the block size~\citep{pmlr-v235-cai24b,chen2026dflash,liu2026dartdiffusioninspiredspeculativedecoding,li2025diffuspecunlockingdiffusionlanguage,sandler2025specdiff2scalingdiffusiondrafter}.
This allows substantially larger blocks (e.g., $\gamma{=}16$) without proportionally increasing latency.

Among them, DFlash~\citep{chen2026dflash} is a state-of-the-art parallel drafter, which conditions its draft model on rich context features extracted from the target model~(KV injection).
During prefill, hidden states from a set of target layers $\{l_1, \ldots, l_m\}$ are concatenated and projected into the draft hidden space:
\begin{equation}
H_{\text{ctx}} = \mathrm{RMSNorm}\bigl(W_c\,[H^{(l_1)};\, \ldots;\, H^{(l_m)}]\bigr),
\label{eq:ctx_feature}
\end{equation}
where $W_c \in \RR^{d \times md}$ is a shared projection.
These context features are injected into every draft layer by concatenating them with the draft block representations along the sequence dimension of keys and values:
\begin{equation}
K_i = [W_i^K H_{\text{ctx}};\; W_i^K H_d], \quad V_i = [W_i^V H_{\text{ctx}};\; W_i^V H_d].
\label{eq:kv_injection}
\end{equation}
All positions within a block attend bidirectionally to each other and to the injected target context.

The draft model shares the target model's embedding layer and language modeling head (both frozen).
It takes as input the embedding of an anchor token\footnote{We use the terms \textit{anchor token} and \textit{bonus token} interchangeably in this paper to denote the final token generated by the target model in the previous decoding round.} followed by $\gamma$ mask token embeddings, and produces logits for all mask positions in a single forward pass.
Since drafting requires only a single forward pass regardless of block size, DFlash can afford deeper architectures and larger blocks than autoregressive drafters under the same latency budget.
\section{Architecture}
\label{sec:dspark}
\begin{figure}[t!]
    \centering
    \includegraphics[width=0.9\linewidth]{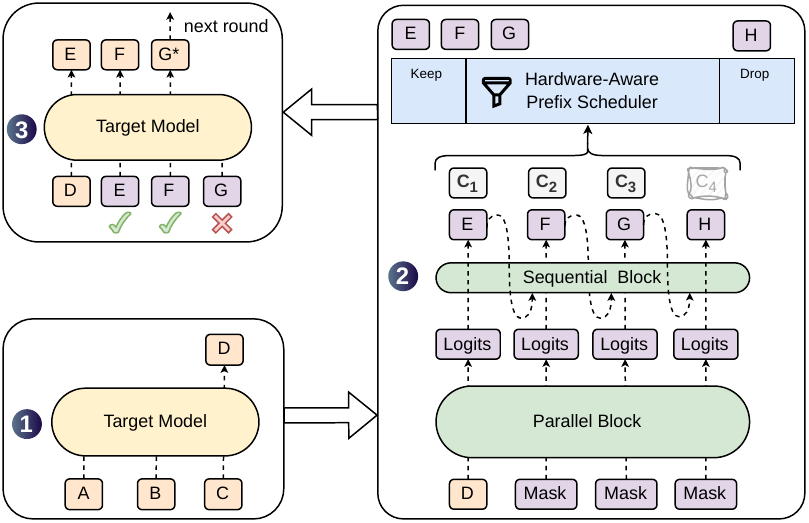}
    \caption{
        \textbf{The DSpark architecture and decoding cycle.} Given prompt tokens \fbox{ABC}, the target model executes one step to generate the next token \fbox{D}, which serves as the anchor for the drafting phase. Using \fbox{D} as the input, DSpark employs a heavy parallel backbone and a lightweight sequential head to generate draft tokens \fbox{EFGH} along with their corresponding confidence scores \fbox{$c_1$--$c_4$}. The Hardware-Aware Prefix Scheduler then evaluates these scores to retain the prefix \fbox{EFG} and drop the low-confidence token \fbox{H}. 
        Finally, the target model verifies the scheduled prefix in parallel. As illustrated, \fbox{E} and \fbox{F} are accepted while \fbox{G} is rejected, prompting the model to generate a corrected token \fbox{G$^*$} to complete the current round.
    }
    \label{fig:model_arch}
\end{figure}

The overview of DSpark is shown in \autoref{fig:model_arch}. Recall from \autoref{eq:latency} that the per-token latency of speculative decoding is $L = (T_{\text{draft}} + T_{\text{verify}})/\tau$.
Autoregressive drafters achieve high $\tau$ but pay $T_{\text{draft}} \propto \gamma$; parallel drafters collapse $T_{\text{draft}}$ to a single pass but sacrifice $\tau$ because each position is predicted independently.
Meanwhile, fixed-length verification wastes $T_{\text{verify}}$ on low-confidence suffix tokens that are almost certain to be rejected.
DSpark addresses these limitations with two complementary components:

\begin{itemize}[leftmargin=*,itemsep=2pt,topsep=4pt]
\item \textbf{Semi-autoregressive generation} (\autoref{sec:semi_ar}). A parallel backbone handles the bulk of draft computation, which keeps $T_{\text{draft}}$ nearly independent of $\gamma$. A lightweight sequential block then injects dependency among draft tokens, improving $\tau$ at minimal additional latency.

\item \textbf{Confidence-scheduled verification} (\autoref{sec:confidence}). A confidence head estimates per-position acceptance probabilities, and a hardware-aware scheduler uses these estimates to prune low-confidence suffix tokens, cutting unnecessary verification compute.
\end{itemize}

\subsection{Semi-Autoregressive Generation}
\label{sec:semi_ar}

A parallel drafter produces all $\gamma$ draft logits in one forward pass, so each prediction cannot condition on tokens sampled elsewhere in the block.
When the context admits multiple plausible continuations, e.g., ``of course'' and ``no problem'', a parallel drafter may produce incoherent combinations such as ``of problem'' or ``no course'', because each position marginalizes over all possible predecessors rather than conditioning on the one actually sampled~\citep{gu2018nonautoregressive,pmlr-v162-huang22k}.
Acceptance rate thus decays rapidly along the block, wasting both draft and verification compute.
We therefore adopt a \textbf{semi-autoregressive} structure that splits draft generation into two stages:

\paragraph{Parallel stage.}
A parallel backbone (in our instantiation, DFlash~\citep{chen2026dflash}) runs a single forward pass over the entire block, producing hidden states $h_1, \ldots, h_{\gamma}$ and base logits $U_1, \ldots, U_{\gamma}$.
We make only a minor modification to the original DFlash backbone: instead of feeding an anchor token plus $\gamma$ mask tokens and predicting only the mask positions, we treat the anchor itself as the first prediction position, so $\gamma$ input tokens (anchor $+$ $\gamma{-}1$ masks) yield $\gamma$ draft logits.
This reduces draft computation while maintaining similar draft quality.

\paragraph{Sequential stage.}
The sequential stage supplements the base logits with a prefix-dependent transition bias
$B_k(x_0, x_{<k}, x_k)$, allowing each draft position to condition on previously sampled
tokens within the block. Rather than defining a globally normalized energy model, the
sequential stage induces a causal block distribution through an autoregressive
factorization:
\begin{equation}
P(X \mid x_0)
=
\prod_{k=1}^{\gamma}
p_k(x_k \mid x_0, x_{<k}),
\qquad
p_k(v \mid x_0, x_{<k})
=
\frac{
\exp\!\left(U_k(v) + B_k(x_0, x_{<k}, v)\right)
}{
\sum_{u \in \mathcal{V}}
\exp\!\left(U_k(u) + B_k(x_0, x_{<k}, u)\right)
}.
\label{eq:sequential-factorization}
\end{equation}
Here, $x_0$ denotes the anchor token from the previous verification cycle, $U_k$ is the
base logit vector produced by the parallel backbone at position $k$, and $\mathcal{V}$ is the
vocabulary. At inference time, the sequential block samples left to right according to
$p_k(\cdot \mid x_0, x_{<k})$.
Because this sampling process is inherently sequential, the block must be computationally
lightweight ($T_{\text{sequential}} \ll T_{\text{parallel}}$) so that the overall draft latency
remains dominated by the parallel stage.
We describe two instantiations of the sequential block below.

\begin{itemize}
    \setlength{\itemsep}{10pt}
    \item \textbf{Markov head.} The simplest instantiation restricts $B_k$ to depend only on the immediately preceding token, reducing it to a first-order transition $B(x_{k-1}, x_k)$.
    In principle this is a full $V \times V$ matrix $B$; we approximate it with a low-rank factorization $B = W_1 W_2$, where $W_1 \in \RR^{V \times r}$ and $W_2 \in \RR^{r \times V}$.
    Given the preceding token $x_{k-1}$, the transition bias for position $k$ is:
    \begin{equation}
    B(x_{k-1},\, \cdot\,) = W_1[x_{k-1}] \, W_2 \;\in\; \RR^{V},
    \label{eq:markov}
    \end{equation}
    where $W_1$ serves as an embedding lookup table and $W_2$ as a logit projection.
    The low-rank factorization ($r{=}256$ by default) keeps both storage and per-step compute small, making the sequential loop efficient even for large vocabularies.
    Returning to the earlier example: once position 1 samples ``of'', the Markov head boosts ``course'' and suppresses ``problem'' at position 2, which mitigates the cross-mode collision.

    \item \textbf{RNN head.} The Markov head is memoryless beyond one step—position $k$ cannot access tokens before $x_{k-1}$.
    The RNN head relaxes this by maintaining a recurrent state $s_k$ that accumulates the full prefix history within a block.
    At each step, the module concatenates the current state $s_{k-1} \in \RR^{r}$, the previous token embedding $W_1[x_{k-1}] \in \RR^{r}$, and the backbone hidden $h_k \in \RR^{d}$ into an input vector $z_k = [s_{k-1};\, W_1[x_{k-1}];\, h_k] \in \RR^{2r+d}$, then applies a single gated update:
    \begin{equation}
    \begin{aligned}
    s_k = \sigma(W_g\, z_k)& \odot s_{k-1} \;+\; \bigl(1 - \sigma(W_g\, z_k)\bigr) \odot \tanh(W_c\, z_k), \label{eq:rnn}\\
    &B_k(x_{<k},\, \cdot\,) = W_2^\top\, \tanh(W_o\, z_k),
    \end{aligned}
    \end{equation}
    where $W_g, W_c, W_o \in \RR^{r\times(2r+d)}$ are jointly parameterized by a single linear projection that is split into gate, candidate, and output components.
    The state $s_0$ is initialized to zero.
\end{itemize}

\subsection{Confidence-Scheduled Verification}  
\label{sec:confidence}  

The semi-autoregressive architecture enables DSpark to generate large draft blocks efficiently. However, producing more draft tokens does not automatically translate to higher end-to-end speedups. Indiscriminately verifying the full draft block can actually degrade overall system throughput, especially in high-concurrency scenarios~\citep{liu2024turbospec,hu2026echo}.

This performance bottleneck stems from two interacting factors. 
First, on the data side, draft acceptance rates inherently vary across domains: structured text like code naturally yields high acceptance, whereas open-ended chat has significantly lower acceptance~\citep{xia-etal-2024-unlocking,abramovich2026speed}.
Second, on the system side, the actual cost of verifying an extra token depends strictly on the engine load. Under light system load, an extra verification incurs minimal penalty even if rejected. However, under high-concurrency deployments, every unnecessary verification occupies target model batch capacity that could otherwise serve other active requests~\citep{liu2024optimizing, wu-etal-2025-tetris}.

Therefore, fully unlocking the potential of large draft blocks requires a unified mechanism that routes target model compute only toward tokens with a positive expected return. DSpark achieves this by coupling a \textbf{confidence head} (\autoref{sec:conf_head}) that predicts prefix survival probabilities, with a \textbf{hardware-aware prefix scheduler} (\autoref{sec:scheduler}) that dynamically determines the optimal verification lengths based on current system load.

\subsubsection{Confidence Head}
\label{sec:conf_head}
Drawing inspiration from~\citet{huang2024specdec++,wang2025end}, the confidence head outputs a scalar   $c_k \in (0, 1)$ for each draft position $k$. Crucially, $c_k$ models the \textit{conditional} probability that the draft token at position $k$ will survive target verification, given that all preceding tokens in the block have been accepted.  
The architecture features a lightweight linear projection followed by a sigmoid function:  
\begin{equation}  
c_k = \sigma\bigl(w^\top [h_k;\, W_1[x_{k-1}]]\bigr),  
\label{eq:confidence}  
\end{equation}  
where $h_k$ is the hidden state of the backbone and $W_1[x_{k-1}]$ is the Markov Embedding from the previous draft token. We supervise $c_k$ using the analytical acceptance rate per-step $c_k^*$. This rate is determined by the total variation distance between the draft distribution $p_k^d$ and the target distribution $p_k^t$:  
\begin{equation}  
c_k^* = 1 - \tfrac{1}{2}\|p_k^d - p_k^t\|_1.  
\label{eq:accept_rate}  
\end{equation}

\paragraph{Post-hoc Calibration.}  
Unlike threshold-based verification heuristics~\citep{huang2024specdec++,li-etal-2024-eagle,zhang2026pacer}, which only require confidence scores to correctly rank draft token qualities, our hardware-aware scheduling approach (detailed in \autoref{sec:scheduler}) precisely requires the absolute magnitudes of the cumulative acceptance probabilities to compute the expected acceptance length $\tau$. Because neural confidence estimates are often overconfident~\citep{guo2017calibration,ovadia2019can}, using the raw confidence scores directly would distort the throughput estimation, leading to suboptimal scheduling.  
  
To address this, we introduce \textbf{Sequential Temperature Scaling (STS)}. Because each $c_i$ models a conditional probability, the chain rule dictates that the joint probability of a draft prefix being accepted factorizes into the cumulative product $\prod_{i \leq k} c_i$. Using a held-out validation set, STS calibrates this joint probability consecutively from left to right. Specifically, at each position $k \in \{1,\dots,\gamma\}$, we perform a simple 1D grid search to find the optimal temperature scalar that minimizes the Expected Calibration Error (ECE)~\citep{naeini2015obtaining} of the cumulative product, keeping the already-calibrated scores of all preceding positions fixed. Crucially, temperature scaling is an order-preserving transformation: it rectifies the predicted probabilities to match empirical acceptance rates without disrupting the relative draft token rankings learned by the confidence head.

\subsubsection{Hardware-Aware Prefix Scheduler}  
\label{sec:scheduler}

\begin{algorithm}[t]
\caption{Hardware-Aware Prefix Scheduler}    
\label{alg:prefix-scheduler}    
\begin{algorithmic}[1]    
\REQUIRE Active requests $r \in \{1,\dots,R\}$; confidence sequence $c_{r,1},\dots,c_{r,\gamma}$ per request; profiled step curve $\text{SPS}(B)$    
\ENSURE Selected per-request prefix lengths $\ell^*_1,\dots,\ell^*_R$    
\FOR{$r = 1$ \TO $R$}    
  \STATE Compute prefix survival probabilities: $a_{r,j} \gets \prod_{i \leq j} c_{r,i}$ for $j = 1,\dots,\gamma$    
\ENDFOR    
\STATE Construct candidate space $\mathcal{E} \gets \{(r, j) \mid a_{r,j} > 0\}$ and sort descending by $a_{r,j}$    
\STATE Initialize states: $\ell_r \gets 0$ for all $r$; Batch size $B \gets R$; Expected accepts $\tau^* \gets R$    
\STATE Initialize tracking: $\Theta_{\text{best}} \gets R \cdot \text{SPS}(R)$; Selected lengths $\ell^*_r \gets 0$ for all $r$    
\FOR{each $(r, j) \in \mathcal{E}$ in sorted order}    
  \STATE $\ell_r \gets j$; $B \gets B + 1$; $\tau^* \gets \tau^* + a_{r,j}$    
  \STATE Current throughput $\Theta \gets \tau^* \cdot \text{SPS}(B)$    
  \IF{$\Theta > \Theta_{\text{best}}$}    
    \STATE $\Theta_{\text{best}} \gets \Theta$; Update selected lengths $\ell^*_r \gets \ell_r$
  \ELSE
    \STATE \textbf{break}
  \ENDIF    
\ENDFOR    
\RETURN $(\ell^*_1,\dots,\ell^*_R)$ achieving $\Theta_{\text{best}}$    
\end{algorithmic}    
\end{algorithm}

Prior methods~\citep{huang2024specdec++,li-etal-2024-eagle} typically apply a static threshold to confidence scores to determine verification length. While effective under isolated, single-request assumptions, static thresholds can be suboptimal in high-concurrency production systems, where the utility of verifying a draft token depends heavily on the current system load.  
  
To address this, we formulate verification length selection as a global throughput maximization problem~(\autoref{alg:prefix-scheduler}). Consider a batch of $R$ active requests. For request $r$, let $c_{r,1},\dots,c_{r,\gamma}$ be the per-position confidence estimates, and let $\ell_r \in \{0,\dots,\gamma\}$ denote the scheduled verification length. Because speculative decoding dynamically accepts draft tokens only as a continuous prefix, the survival probability of a token at position $j$ is the cumulative product $a_{r,j} = \prod_{i \leq j} c_{r,i}$.  
  
In a single verification step, the total batch size (measured in tokens) sent to the target model is $B = \sum_{r=1}^{R} (1 + \ell_r)$, and the expected number of successfully accepted tokens is $\tau = \sum_{r=1}^{R} \bigl(1 + \sum_{j=1}^{\ell_r} a_{r,j}\bigr)$. 
Under a simplifying assumption\footnote{In practical serving scenarios, average context lengths remain well below extremes (e.g., 1M tokens), making their impact on decode latency marginal for highly optimized architectures like DeepSeek-V4. Moreover, in prefill-decode disaggregated deployments, decode load balancers keep both request counts and total context lengths roughly balanced across data-parallel (DP) ranks. This effectively amortizes sequence-length variance, allowing us to safely assume that engine throughput depends predominantly on the verification batch size $B$.}, let $\text{SPS}(B)$ denote the engine throughput, measured in steps per second, for a given forward-pass batch size $B$. Crucially, this capacity curve is profiled once during engine initialization and stored as a lightweight cost table. Our scheduler then aims to maximize the expected system-wide token throughput $\Theta = \tau \cdot \text{SPS}(B)$ by dynamically selecting verification lengths $\ell_1,\dots,\ell_R$.
  
Although finding the global maximum of $\Theta$ appears to be a combinatorial search, the objective structure allows for an efficient greedy solution. Because $a_{r,j}$ is monotonically non-increasing with respect to $j$ (i.e., $a_{r, j} \le a_{r, j-1}$), the marginal gain in expected accepted tokens for extending request $r$'s verification length from $j-1$ to $j$ is exactly $a_{r,j}$. This monotonicity ensures that sorting candidate tokens globally by $a_{r,j}$ naturally respects intra-block prefix dependencies. Consequently, if the total verification batch size $B$ were fixed, the optimal allocation $\{\ell_r\}$ would be determined by greedily selecting the draft tokens with the highest survival probabilities from the global pool of all $\{a_{r,j}\}$.

Building on this insight, the optimization can be evaluated along this greedy admission path. We first globally sort all valid prefix extensions in descending order of survival probability. To dynamically determine the optimal target batch size $B$, we incrementally admit tokens from this sorted pool, updating the expected throughput $\Theta$ via an lookup from the cost table. 

Lossless speculative decoding strictly requires the \textit{non-anticipating property}: admission decisions must not depend on future candidate tokens~\citep{pmlr-v202-leviathan23a,chen2023accelerating}. Because our confidence head relies on the Markov feature of the previously sampled token, computing the next survival probability $a_{r,k+1}$ explicitly requires the instantiated candidate $x_{r,k}$. A retrospective global search would thus inadvertently leak $x_{r,k}$ into the admission decision for step $k$, introducing selection bias~(we provide a concrete counterexample demonstrating this theoretical violation in \autoref{app:counterexample}).

To enforce strict causality, the scheduler (\autoref{alg:prefix-scheduler}) employs an early-stopping mechanism. By breaking the greedy search immediately when the throughput drops ($\Theta \le \Theta_{\text{best}}$), the truncation decision relies solely on the prefix processed up to that exact step. This isolates the admission event from future tokens, ensuring exact target-distribution recovery. Note that this stepwise early-stopping yields the global maximum throughput if and only if the objective $\Theta$ is unimodal, which implicitly assumes a smoothly decaying hardware capacity curve. We address the engineering adaptations required for real-world, non-smooth $\text{SPS}$ characteristics and asynchronous system pipelines in \autoref{sec:scheduler_in_practice}.

\subsection{Training}
\label{sec:training}

During training, we randomly sample multiple anchor positions from each target sequence to form $\gamma$-token blocks as training data.
The target model is frozen throughout training; the draft model shares its embedding layer and language modeling head and keeps them frozen, updating only the backbone drafter, sequential block, and confidence head.

The training objective consists of three terms: a cross-entropy loss $\Ll_{\text{ce}}$, a distribution-matching loss $\Ll_{\text{tv}}$, and a confidence loss $\Ll_{\text{conf}}$.
All three are position-weighted by $w_k = \exp(-(k{-}1)/\gamma)$~\citep{chen2026dflash}, which emphasizes earlier block positions that contribute more to the expected acceptance length under prefix-based verification.
The cross-entropy loss $\Ll_{\text{ce}}$ trains the drafter to predict the correct next token:
\begin{equation}
\Ll_{\text{ce}} = -\sum_{k=1}^{\gamma} w_k \log p^d_{k}(x_k^*),
\end{equation}
where $x_k^*$ is the ground-truth token and $p^d_k$ is the draft distribution.
The distribution-matching loss $\Ll_{\text{tv}}$ penalizes the total variation distance between the draft and target distributions:
\begin{equation}
\Ll_{\text{tv}} = \sum_{k=1}^{\gamma} w_k \|p_k^d - p_k^t\|_1.
\end{equation}
Since the total variation distance is a direct proxy for the acceptance rate: the per-step acceptance probability equals $1 - \frac{1}{2}\|p^d - p^t\|_1$~\citep{pmlr-v202-leviathan23a}, minimizing $\Ll_{\text{tv}}$ directly maximizes the expected acceptance rate.

The confidence loss $\Ll_{\text{conf}}$ is a binary cross-entropy that trains the confidence head to predict the soft acceptance label $c_k^*$ from \autoref{eq:accept_rate}:
\begin{equation}
\Ll_{\text{conf}} = -\sum_{k=1}^{\gamma} w_k \bigl[c_k^* \log c_k + (1 - c_k^*) \log(1 - c_k)\bigr].
\end{equation}
The overall objective is a weighted combination of the three terms~(with default weights $\alpha_{\text{ce}} = 0.1$, $\alpha_{\text{tv}} = 0.9$, $\alpha_{\text{conf}} = 1.0$):
\begin{equation}
\Ll = \alpha_{\text{ce}}\,\Ll_{\text{ce}} + \alpha_{\text{tv}}\,\Ll_{\text{tv}} + \alpha_{\text{conf}}\,\Ll_{\text{conf}}
\label{eq:loss}
\end{equation}

\section{Experiments}
\label{sec:experiments}
In this section, we validate the draft quality of DSpark using offline benchmarks and report the effectiveness of confidence scheduler under online production traffic in \autoref{sec:real_world_deployment}. The experimental setup is described in \autoref{sec:exp_setup}, main results in \autoref{sec:exp_main_results}, and additional analyses are included in \autoref{sec:exp_analysis}.
\subsection{Experimental Setup}
\label{sec:exp_setup}

\paragraph{Target and draft models.}
We evaluate DSpark on four target models spanning different scales and model families: Qwen3-\{4B, 8B, 14B\}~\citep{yang2025qwen3}, and Gemma4-12B~\citep{gemma4}. 
For draft models, we compare DSpark with two representative drafters: DFlash~\citep{chen2026dflash}, a state-of-the-art parallel drafter, and Eagle3~\citep{li2026eagle}, an autoregressive drafter based on Training-Time Test (TTT).
For strict and fair comparison, we retrain all drafters in the same \href{https://github.com/deepseek-ai/DeepSpec}{training framework} and on the same data\footnote{To facilitate future research, we release all \href{https://huggingface.co/collections/deepseek-ai/deepspec}{checkpoints} we trained, including Eagle3, DFlash and DSpark.}.
We align Eagle3's TTT horizon~(7) with the block size~(7) used by DFlash and DSpark, and we use the same target-model feature layers for all drafters.
For the number of draft model layers, we set 1 for Eagle3 and 5 for DSpark and DFlash~\citep{chen2026dflash}. 
Unless otherwise stated, DSpark denotes the Markov-head variant; we study the RNN-head variant in \autoref{sec:exp_depth_width}.

\paragraph{Training data.}
We use \href{https://huggingface.co/datasets/mlabonne/open-perfectblend}{Open-PerfectBlend}, an open-sourced version of PerfectBlend~\citep{xu2024perfect} consisting of 1.3 million samples. It is a general-purpose instruction dataset containing chat (17.6\%), math (39.4\%), code (38.9\%), and instruction-following data (4.1\%).
We only use the prompts from Open-PerfectBlend; responses are regenerated by each target model with recommended sampling parameters. Each drafter is trained for 10 epochs to ensure full convergence.
For data generation and evaluation, we adopt the non-thinking mode. 

\paragraph{Evaluation protocol.}
We evaluate the performance of different algorithms on three domains:
\begin{enumerate}
    \item \textbf{Mathematical Reasoning}, including GSM8K~\citep{cobbe2021gsm8k}, MATH500~\citep{lightman2023lets} and AIME25~\citep{aime25}.
    \item \textbf{Code Generation}, including MBPP~\citep{austin2021program}, HumanEval~\citep{chen2021codex} and Live-CodeBench~\citep{jain2025livecodebench}.
    \item \textbf{Daily Chat}, including MT-Bench~\citep{zheng2023judging}, Alpaca~\citep{alpaca} and Arena-Hard~\citep{arenahard2024,li2024crowdsourced}.
\end{enumerate}
For all benchmarks, we use standard speculative decoding~\citep{pmlr-v202-leviathan23a,chen2023accelerating} with the sampling temperature set to $1.0$. We report the accepted length~($\tau$) per decoding round\footnote{For clarity, unless otherwise stated, all reported metrics for accepted length and acceptance rate include the target-generated bonus token.}. For all drafters, we use chain-based drafting.

\begin{table}[thbp]
\centering
\caption{\textbf{Main speculative decoding results.} We report accepted length~($\tau$) per decoding round~(higher is better) for different target models and domains. \textbf{Bold} marks the best results.}
\label{tab:main_exp}
\resizebox{0.99\textwidth}{!}{
\begin{tabular}{@{}ll ccc ccc ccc@{}}
\toprule
\multirow{2}{*}{Target} & \multirow{2}{*}{Drafter} & \multicolumn{3}{c}{Math} & \multicolumn{3}{c}{Code} & \multicolumn{3}{c}{Chat} \\
\cmidrule(lr){3-5} \cmidrule(lr){6-8} \cmidrule(lr){9-11}
 & & GSM8K & MATH & AIME25 & MBPP & HumanEval & LCB & MT-Bench & Alpaca & Arena-Hard \\
\midrule
\multirow{3}{*}{Qwen3-4B}   & Eagle3 & 5.14 & 4.62 & 3.92 & 3.69 & 4.16 & 3.77 & 2.39 & 2.26 & 2.55 \\
                            & DFlash & 5.40 & 4.85 & 4.15 & 4.40 & 4.74 & 4.18 & 3.07 & 2.96 & 2.83 \\
                            & DSpark & \textbf{6.11} & \textbf{5.70} & \textbf{4.89} & \textbf{5.13} & \textbf{5.38} & \textbf{4.86} & \textbf{3.64} & \textbf{3.54} & \textbf{3.29} \\
\midrule
\multirow{3}{*}{Qwen3-8B}   & Eagle3 & 5.30 & 4.77 & 3.91 & 3.96 & 4.33 & 4.17 & 2.66 & 2.54 & 2.54 \\
                            & DFlash & 5.33 & 4.91 & 4.07 & 4.36 & 4.64 & 4.39 & 3.11 & 2.98 & 2.81 \\
                            & DSpark & \textbf{6.17} & \textbf{5.78} & \textbf{5.01} & \textbf{5.16} & \textbf{5.52} & \textbf{5.17} & \textbf{3.72} & \textbf{3.58} & \textbf{3.21} \\
\midrule
\multirow{3}{*}{Qwen3-14B}   & Eagle3 & 5.24 & 4.60 & 3.71 & 3.81 & 4.14 & 4.01 & 2.62 & 2.47 & 2.48 \\
                            & DFlash & 5.41 & 4.84 & 3.98 & 4.44 & 4.59 & 4.33 & 3.10 & 2.94 & 2.72 \\
                            & DSpark & \textbf{6.21} & \textbf{5.74} & \textbf{4.94} & \textbf{5.26} & \textbf{5.43} & \textbf{5.02} & \textbf{3.70} & \textbf{3.58} & \textbf{3.13} \\
\midrule
\multirow{3}{*}{Gemma4-12B}   & Eagle3 & 5.87 & 5.46 & 4.83 & 4.72 & 5.37 & 4.16 & 3.19 & 3.06 & 2.72 \\
                            & DFlash & 5.45 & 5.04 & 4.22 & 4.39 & 4.95 & 3.70 & 2.98 & 2.84 & 2.59 \\
                            & DSpark & \textbf{6.05} & \textbf{5.78} & \textbf{5.12} & \textbf{5.11} & \textbf{5.64} & \textbf{4.51} & \textbf{3.49} & \textbf{3.35} & \textbf{2.92} \\
\bottomrule
\end{tabular}
}
\end{table}

\subsection{Experimental Results}
\label{sec:exp_main_results}

To isolate the raw draft quality from system-level scheduling policies, our offline evaluation disables the confidence scheduler, forcing all drafters to propose a fixed block of tokens. The main results, measured by the average accepted length ($\tau$) per round, are reported in \autoref{tab:main_exp}.

DSpark consistently outperforms both the autoregressive baseline (Eagle3) and the parallel baseline (DFlash) across all evaluated target models and benchmark domains. 
Specifically, across the Qwen3-4B, 8B, and 14B models, DSpark improves the macro-average accepted length over Eagle3 by 30.9\%, 26.7\%, and 30.0\%, respectively. Similarly, compared to DFlash, DSpark yields relative improvements of 16.3\%, 18.4\%, and 18.3\% across the three scales.
Crucially, this advantage generalizes across model families, as demonstrated by the consistent performance gains on the Gemma4-12B target.

Beyond the average improvements, \autoref{tab:main_exp} reveals a strong domain effect: the accepted length is naturally higher on structured tasks (e.g., 5.57 on math and 5.12 on code for Qwen3-4B) than on open-ended chat (3.49). 
This inherent variance in data predictability means a static verification length often wastes compute on trailing tokens that are highly likely to be rejected. This directly motivates our confidence-scheduled verification, which dynamically prunes the draft block based on expected acceptance.

\subsection{Experimental Analysis}
\label{sec:exp_analysis}

\subsubsection{Why Can Parallel Generation Outperform Autoregression?} 
\label{sec:parallel_vs_ar_analysis} 

\begin{figure}[t]   
    \centering  
    \includegraphics[width=0.95\linewidth]{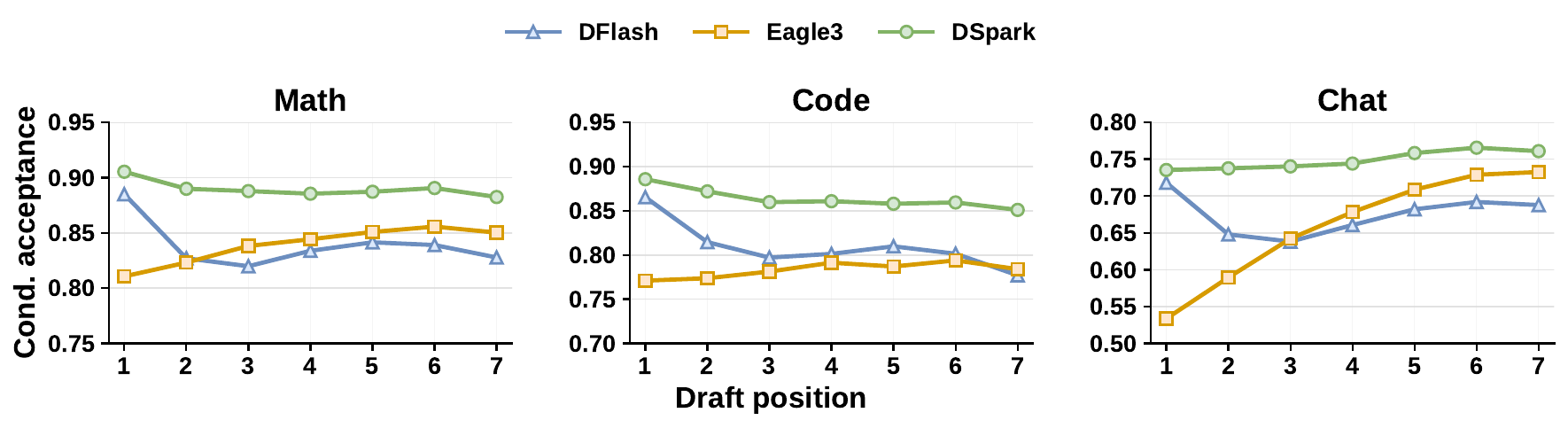}   
    \caption{\textbf{Position-wise conditional acceptance.} We report the empirical conditional acceptance rate for each draft position, averaged across benchmarks within each domain using the Qwen3-4B target model. Unlike standard prefix survival, this metric isolates the baseline predictive quality at position $k$ by removing the penalty of previous rejections. Notice that the autoregressive drafter (Eagle3) remains stable or trends upward, while the parallel drafter (DFlash) suffers suffix decay.}   
    \label{fig:position_cond_accept}   
\end{figure}   

\autoref{tab:main_exp} presents a counter-intuitive observation: the parallel drafter (DFlash) and the semi-autoregressive drafter (DSpark) often yield longer accepted lengths than the fully autoregressive drafter (Eagle3).
This finding contrasts with the standard expectation that step-by-step autoregression produces higher-quality sequences than parallel models~\citep{ren2020study, israel2026accelerating,zheng2025maskeddiffusionmodelssecretly}.

To analyze this behavior, we examine performance beyond the macro-level accepted length. Using the Qwen3-4B target model and the benchmark sets described in \autoref{sec:exp_setup}, we introduce \textit{position-wise conditional acceptance} tracked during actual speculative decoding rollouts. Specifically, for a given draft position $k$, the evaluation denominator counts only the instances where the target model successfully verifies and accepts all preceding draft tokens from $1$ to $k-1$. The metric then calculates the proportion of these valid instances where the token at position $k$ is also accepted. This approach ensures that the evaluation of position $k$ is not penalized by earlier prefix errors, revealing the underlying predictive quality at each specific step. \autoref{fig:position_cond_accept} details these measurements, demonstrating clear behavioral differences across the architectures.

\paragraph{The Capacity Advantage at Position 1.} 
At the first draft position, both architectures predict the next token based solely on the target context. The performance divergence here stems strictly from architectural capacity: autoregressive models like Eagle3 are constrained to shallow networks due to their $O(\gamma)$ latency, whereas $O(1)$ parallel drafters can afford much deeper networks. This structural gap yields a substantial accuracy margin at position 1, with DFlash starting noticeably higher than Eagle3 (e.g., 0.88 vs.\ 0.81 on Math, and 0.72 vs.\ 0.53 on Chat). Because speculative decoding operates as a strict prefix-matching survival process, the first token carries the highest leverage—a rejection here immediately invalidates the entire block. Consequently, this initial capacity advantage disproportionately boosts the final accepted length, explaining why parallel drafters ultimately outperform autoregressive ones globally despite rapid acceptance decay at later positions.

\paragraph{The Limitation of Independence at Later Positions.} 
Examining the tail of the curves (positions 2 through 7) exposes the inherent limitation of independent parallel generation. As earlier tokens lock in a specific semantic path, subsequent tokens naturally become more predictable. Autoregressive models like Eagle3 effectively leverage this conditional certainty, maintaining or even increasing conditional acceptance deeper into the block (e.g., from 0.53 to 0.74 on Chat). In contrast, DFlash suffers from rapid acceptance decay, dropping from 0.87 to 0.78 on Code and 0.72 to 0.63 on Chat. Because each parallel position marginalizes over all possible prior tokens rather than conditioning on an exact sampled prefix, the model frequently proposes inconsistent suffix combinations—a mode known as multi-modal collision~\citep{gu2018nonautoregressive,stern2018blockwiseparalleldecodingdeep}.

\paragraph{Mitigating Suffix Decay with Semi-Autoregression.}
The preceding analysis highlights a clear architectural objective: combining the high capacity of a parallel backbone for the initial token with the dependency modeling of an autoregressive model for subsequent tokens. This directly motivates DSpark's semi-autoregressive design. As shown in \autoref{fig:position_cond_accept}, DSpark inherits the high initial acceptance of the deep parallel drafter (e.g., starting at 0.93 on Math). Simultaneously, its lightweight sequential head mitigates the rapid acceptance decay typical of parallel generation. By resolving this trade-off, DSpark maintains a high and stable conditional acceptance rate throughout the entire draft block.

\subsubsection{A Little Autoregression Goes a Long Way}  
\label{sec:exp_depth_width}  

Building on the insights from \autoref{sec:parallel_vs_ar_analysis}, we explore the architectural design space of DSpark along two dimensions: drafter depth (number of transformer layers) and proposal length (block size $\gamma$).  
Unless otherwise stated, all experiments in this section use Qwen3-4B as the target model and follow the evaluation protocol detailed in \autoref{sec:exp_setup}.

\begin{figure}[t]
    \centering
    \includegraphics[width=0.9\linewidth]{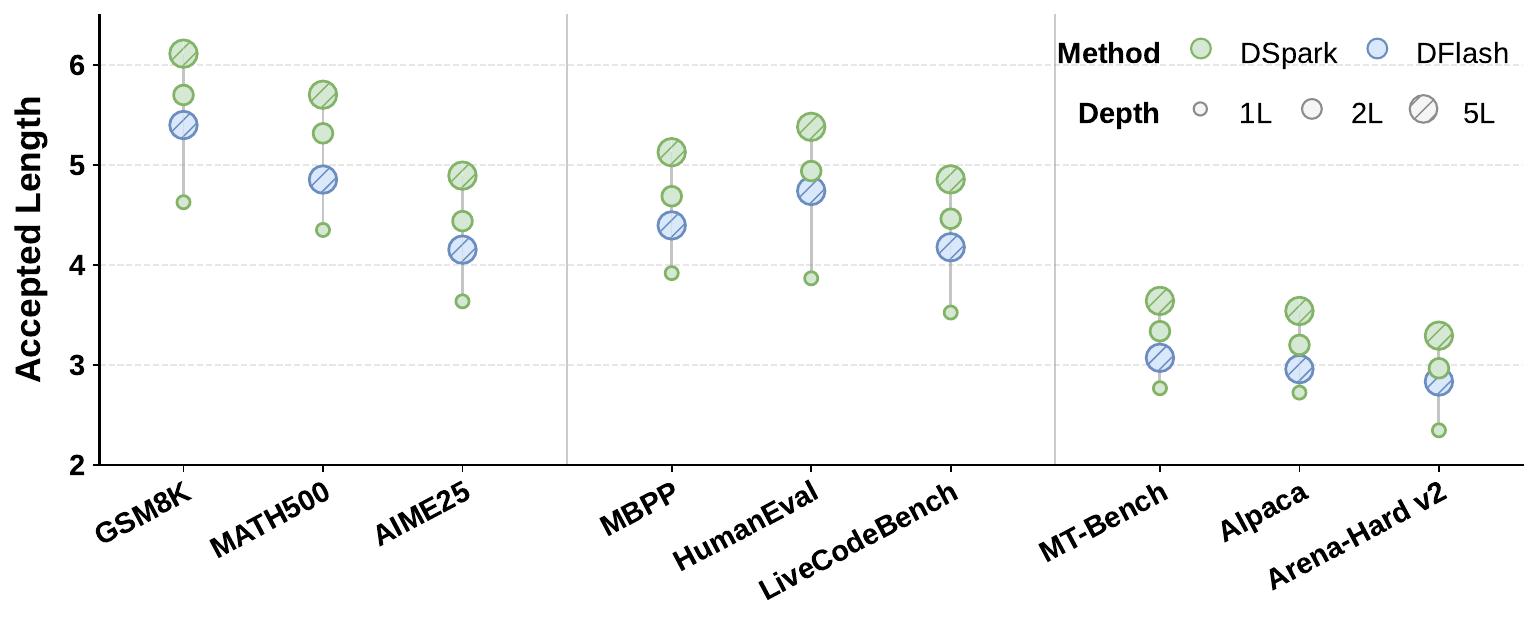}
    \caption{\textbf{Effect of drafter depth.} With proposal length fixed, DSpark's performance improves as drafter layers are added. Notably, a shallow 2-layer DSpark outperforms a deeper 5-layer DFlash baseline, highlighting the parameter efficiency of sequential modeling.}
    \label{fig:layer_comparison}
\end{figure}

\begin{figure}[t]
    \centering
    \includegraphics[width=0.95\linewidth]{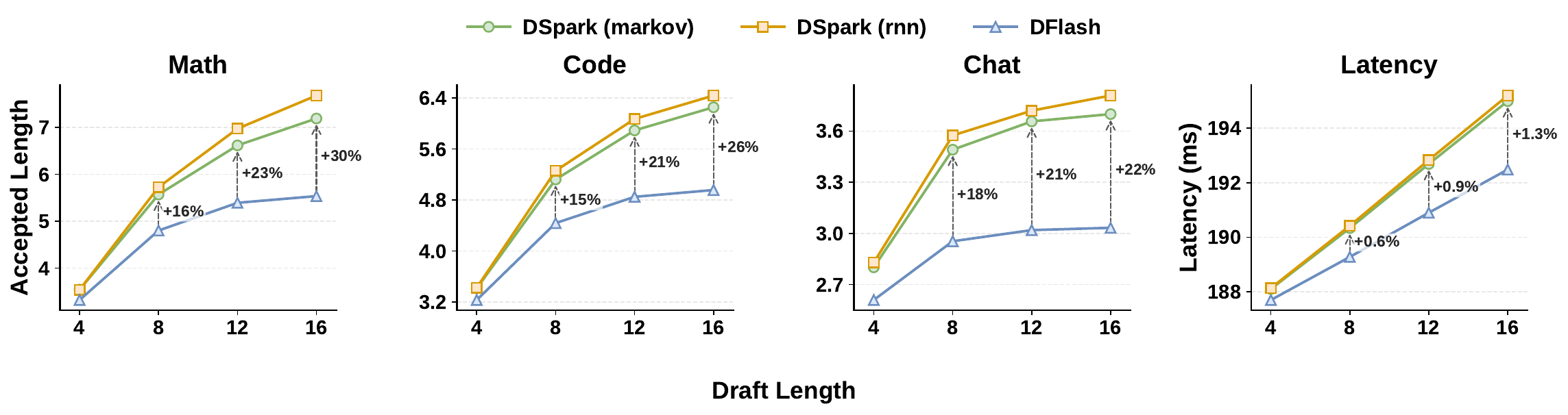}
    \caption{\textbf{Effect of proposal length and latency overhead.} DSpark consistently outperforms DFlash across various block sizes (left three panels). The rightmost panel demonstrates that the sequential head introduces minimal latency overhead during serving.}
    \label{fig:length_comparison}
\end{figure}

\paragraph{Drafter Depth.}  
Increasing the number of transformer layers naturally expands a draft model's predictive capacity. To isolate this effect, we fix the block size to 7 and vary the number of DSpark layers from 1 to 5, comparing it against a 5-layer DFlash baseline. 
\autoref{fig:layer_comparison} aggregates the accepted lengths across the math, code, and chat domains.  
As expected, DSpark's performance improves monotonically with depth, with the steepest marginal gain occurring from one to two layers.  
Notably, a 2-layer DSpark outperforms the 5-layer DFlash baseline across all domains. This demonstrates that injecting local auto-regression via a lightweight sequential head offers a highly favorable accuracy-parameter trade-off, achieving better sequence coherence than simply stacking deeper parallel layers.

\paragraph{Proposal Length.}
Next, we fix the drafter depth to 5 layers and scale the draft length (proposal length $\gamma$ plus one anchor token) across $\{4, 8, 12, 16\}$ to evaluate performance on longer draft blocks. For DSpark, we evaluate both the default Markov head and the RNN head. The first three panels of \autoref{fig:length_comparison} show that DSpark consistently outperforms DFlash at every proposal length. More importantly, the performance gap steadily widens as $\gamma$ increases. Because pure parallel generation (DFlash) suffers from rapid acceptance decay (\autoref{fig:position_cond_accept}), its marginal utility diminishes for long blocks. DSpark mitigates this decay, causing its relative gain over DFlash to grow. For instance, at $\gamma=7$, DSpark improves the accepted length by 16\% on math, 15\% on code, and 18\% on chat; at $\gamma=15$, these gains expand to 30\%, 26\%, and 22\%, respectively. Also, RNN head provides only marginal additional gains over the Markov head, mainly at longer proposal lengths. Given its higher implementation complexity and less favorable deployment properties, we use the Markov head as the default.

\paragraph{Latency Overhead.}  
We quantify the overhead of the sequential generation loop in DSpark. The rightmost panel of \autoref{fig:length_comparison} reports the per-round engine latency—comprising one target verification pass, the parallel draft block forward, and the serial sampling loop—measured at a batch size of 128. To prevent sequence-length bias, the reported latency represents the arithmetic mean across varying context lengths ($\{512, 1024, 2048, 4096\}~ \text{tokens}$). 
Since the target model dominates the verification compute time at this batch size, the sequential block's latency overhead is negligible. Consequently, scaling the draft length from 4 to 16 adds a marginal 0.2\% to 1.3\% to the full-round latency over the DFlash baseline, despite delivering up to a 30\% improvement in accepted length.  

\subsubsection{Verify Smarter, Not Longer: The Role of Confidence Head}
\label{sec:exp_confidence}

While DSpark sustains high acceptance over long draft blocks, verifying the entire proposal remains inefficient~\citep{huang2024specdec++,hu2026echo}. Due to the inherent domain variance noted in \autoref{sec:exp_main_results}, trailing tokens in open-ended chat still face high rejection risks, making blind verification a waste of target compute. To evaluate whether the confidence head can effectively prune these unpromising suffixes, we conduct an offline threshold sweep using Qwen3-4B. We validate the estimator in isolation here, reserving the hardware-aware prefix scheduler (\autoref{sec:scheduler}) for live production evaluation in \autoref{sec:real_world_deployment}.

\begin{figure}[t]
    \centering
    \includegraphics[width=0.95\linewidth]{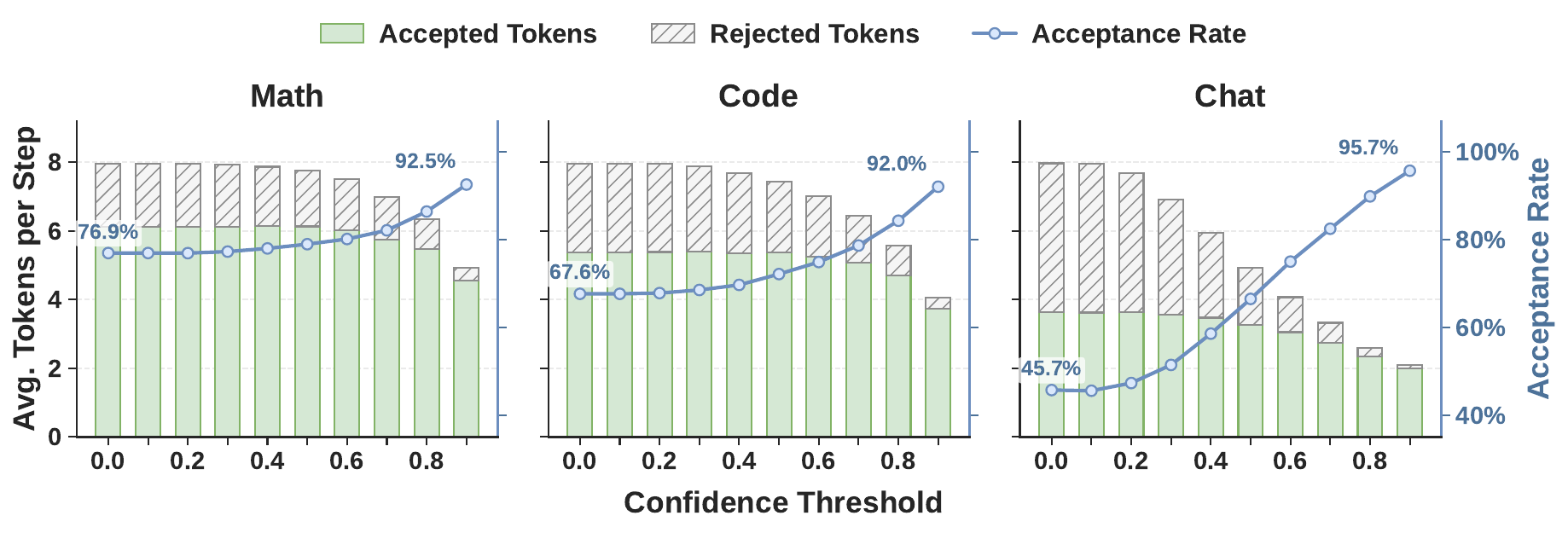}
    \caption{\textbf{Confidence threshold sweep.} A threshold of 0 corresponds to standard fixed-length verification. As the threshold increases, the overall acceptance rate steadily rises because the confidence head effectively prunes tokens that would ultimately be rejected (hashed bars).}
    \label{fig:confidence_head_sweep}
\end{figure}

\begin{figure}[t]
    \centering
    \includegraphics[width=0.95\linewidth]{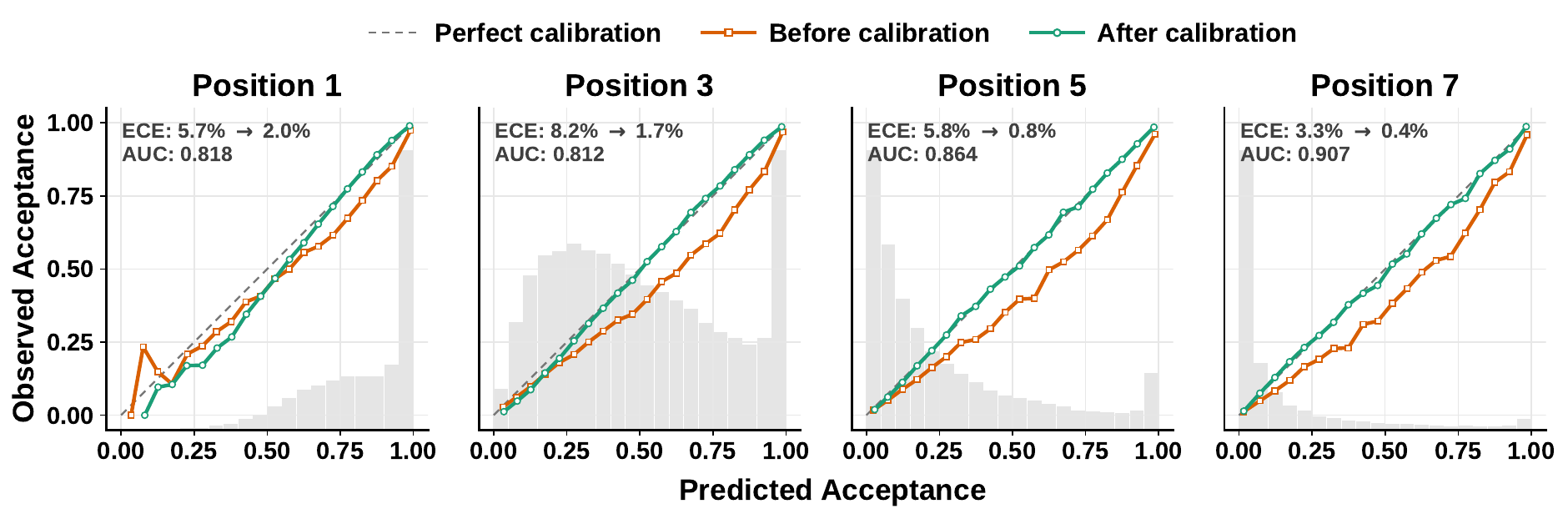}
    \caption{\textbf{The Reliability Diagram on Alpaca Dataset.} While the raw confidence estimator achieves strong discrimination, its predictions are inherently overconfident. Applying post-hoc calibration helps to align the prefix survival probabilities with empirical acceptance rates. The shaded background histogram represents the frequency distribution of sample counts across different confidence bins.}
    \label{fig:confidence_head_calibration}
\end{figure}

\paragraph{Diagnostic: Static Threshold Sweep.} 
\autoref{fig:confidence_head_sweep} plots the average tokens per step (bars) and the overall acceptance rate (line) across confidence thresholds. As the threshold increases, the acceptance rate steadily rises because the estimator filters out tokens that would ultimately be rejected (hashed bars). 
This suggests that the confidence head can identify lower-value suffix tokens and this pruning is most pronounced on chat workloads, where higher-entropy token distributions limit the efficiency of fixed-length verification. In the Chat subplot, raising the threshold significantly reduces rejected tokens, increasing the acceptance rate from 45.7\% to 95.7\%. In contrast, structured tasks (Math and Code) experience milder pruning and retain more draft tokens, with acceptance rates rising from 76.9\% to 92.5\% and 67.6\% to 92.0\%, respectively. 

\paragraph{From Static Thresholds to Calibrated Scheduling.}
While useful for diagnostics, a static threshold is sub-optimal in dynamic serving environments because it ignores system load: verifying low-confidence tokens incurs minimal opportunity cost under low concurrency, but wastes critical batch capacity under high concurrency. This load dependency motivates the hardware-aware prefix scheduler. As formulated in \autoref{sec:confidence}, maximizing system-level throughput requires the confidence model to exhibit both strong predictive discrimination and precise calibration to accurately estimate cumulative survival probabilities. The reliability diagram~(\autoref{fig:confidence_head_calibration}) demonstrates that while the raw model achieves strong discrimination (ROC-AUC~\citep{hanley1982meaning} ranging from 0.81 to 0.90), it is overly confident (ECE 3\%--8\%). Applying post-hoc STS~(\autoref{sec:conf_head}) mitigates this overconfidence, reducing the average ECE to $\sim$1\% and yielding reliable survival estimates.
\section{Real-World Deployment of DSpark}   
\label{sec:real_world_deployment}  

While \autoref{sec:experiments} establishes the algorithmic gains of DSpark on offline benchmarks, deploying it alongside large-scale models like DeepSeek-V4~\citep{deepseekai2026deepseekv4highlyefficientmilliontoken} introduces additional system-level challenges across both training and inference. In this section, we present the end-to-end production pipeline of DSpark. We detail our scalable training mechanisms, the system-level optimizations necessary to deploy the hardware-aware prefix scheduler~(\autoref{sec:scheduler}), and the framework's end-to-end performance under live user traffic.

\subsection{Scalable and Flexible Training}   
\label{sec:prod_training}  
  
The DSpark draft models are co-deployed with the preview versions of DeepSeek-V4-Flash and DeepSeek-V4-Pro~\citep{deepseekai2026deepseekv4highlyefficientmilliontoken}. The parallel backbone comprises three MoE layers~\citep{dai2024deepseekmoe} with mHC~\citep{xie2025mhc} and a sliding window attention of 128. We configure the maximum block size to $\gamma=5$ and utilize the Markov head for sequential modeling. 
Furthermore, the confidence head is trained end-to-end alongside the draft model and subsequently calibrated via STS to provide reliable scheduling signals.
  
Training the draft model requires the target model's output distributions for supervision. Evaluating both models over the full document context incurs substantial memory footprints and inter-worker communication overhead. To address these bottlenecks, we implement two system-level optimizations within our internal training framework~(HAI-LLM)\footnote{\url{https://www.high-flyer.cn/en/blog/hai-llm/}}:

\begin{itemize}[leftmargin=*,itemsep=4pt,topsep=4pt]   
    \item \textbf{Hidden state communication.} Transferring the target model's full-vocabulary logits ($V \approx 10^5$) across parallel workers creates a significant bandwidth bottleneck. Instead, we temporarily cache the target model's forward-pass activations and communicate only the hidden states immediately preceding the language modeling (LM) head. The LM head projection is then executed locally on the draft model's workers only for the sampled target positions. This reduces the per-token communication complexity to $O(d)$, where $d$ is the hidden dimension.
      
    \item \textbf{Anchor-bounded sequence packing.} To decouple the draft model's computational cost from the target model's context length, we sample a fixed number of draft anchors from the training sequence and pack these isolated prediction blocks into dense training batches. We manage this packing via token-level attention indices rather than standard 2D masks. This maintains exact causal masking across multiple independent sequences and anchors, avoiding the computational and memory overhead associated with standard padding.  
\end{itemize}

\subsection{Hardware-Aware Prefix Scheduler in Practice}
\label{sec:scheduler_in_practice}

In \autoref{sec:scheduler}, \autoref{alg:prefix-scheduler} provides a theoretically sound and lossless scheduling mechanism. However, directly deploying this algorithm into a production environment exposes two fundamental conflicts with real-world infrastructure. 
First, the algorithm assumes a smooth, unimodal capacity curve, whereas the true hardware capacity $\text{SPS}(B)$ is inherently discrete, exhibiting a jagged, step-wise degradation~\citep{Yan2020DemystifyingTC}. 
Second, the algorithm requires scheduling of dynamic draft tokens per step, which clashes with continuous CUDA graph replay~\citep{fireworks2023cudagraphs} and Zero-Overhead Scheduling (ZOS)~\citep{zhu2025nanoflowoptimallargelanguage,zheng2024sglangefficientexecutionstructured}. 

To navigate the trade-offs among system compatibility, throughput, and algorithmic correctness, we adapt the scheduler to operate asynchronously. Because ZOS requires the batch size for the next step to be known before the current step completes, synchronous scheduling would inevitably stall the GPU pipeline. Instead, we approximate the upcoming verification capacity using the confidence head outputs from two steps prior. Mechanically, the candidate tokens in the current step are still strictly sorted by their actual, up-to-date cumulative confidence scores; the historical prediction from two steps prior is used solely to determine the dynamic truncation length (i.e., the batch capacity limit $K$). This effectively casts the admission process as a dynamic top-$K$ selection. While approximating the capacity $K$ introduces a slight temporal offset, the selection mechanism is fundamentally rank-preserving: the most confident draft tokens are always prioritized for verification. This adaptation fully hides scheduling latency and ensures seamless ZOS integration.

Building on this asynchronous pipeline, we resolve the hardware utilization bottleneck. To prevent the scheduler from being trapped in local minima by jagged $\text{SPS}$ cliffs, we remove the early-stopping \texttt{break}, enabling an unconstrained global search. Ordinarily, this retrospective search would leak future token information and violate the lossless guarantee~(\autoref{app:counterexample}). However, our ZOS-driven adaptation naturally prevents this. Because the unconstrained search evaluates only historical predictions from two steps prior, the admission decision is isolated from the realization of the current token $x_{r,k}$. The truncation length inherently depends only on information available from two steps prior. Thus, asynchronous design forms a causal barrier, maximizing physical throughput across hardware cliffs while preserving the exact target distribution.

\subsection{High-Throughput and Low-Latency Inference} 
\label{sec:prod_systems}

During decoding, production serving systems must simultaneously optimize two competing objectives: per-request latency and aggregate throughput~\citep{kwon2023efficient,zhong2024distserve,zhao2025insights}. The former governs the quality of service for individual users---a factor increasingly critical in agent-based workloads~\citep{tiwari2026cachewise}---while the latter determines the total number of concurrently served users. Because speculative decoding inevitably incurs wasted verification compute, it inherently navigates this trade-off, trading extra system compute for faster per-request generation. 

In our deployment setting, however, the number of requests processed per step is frequently constrained by resource limits (e.g., fixed KV-cache capacity per request) and the pool of available user traffic (e.g., RL long-tail loads). Consequently, the effective batch size persistently remains well below the GPU's compute-saturating threshold. 
Under this regime, the traditional trade-off simplifies: given a fixed concurrency limit, maximizing per-GPU total token throughput and maximizing the generation speed per user~(\textit{tok/s/user}) become highly correlated objectives rather than competing ones.

To achieve this maximum throughput, the asynchronous scheduler (\autoref{sec:scheduler_in_practice}) actively routes idle compute toward the most promising draft tokens. However, executing this dynamic routing introduces a severe challenge at the physical execution layer: the inference framework must efficiently support variable-length queries within a single batch. Standard decode kernels are heavily optimized for fixed query lengths; naively processing variable-length verified prefixes leads to severe GPU under-utilization due to padding and uneven workload distribution. We resolve this by decoupling physical execution from logical sequence tracking. In our compute kernels, all tokens across different requests are flattened and processed identically as independent elements. The complex intra-sequence dependencies are then strictly conveyed via a marker tensor integrated into our sparse attention implementation. Specifically on the DeepSeek-V4 architecture, only the index-attention and compress kernels require modification to support this variable-length routing, allowing the dynamic scheduler to operate seamlessly without introducing low-level execution overhead.

\subsection{Performance under Live User Traffic}
\label{sec:prod_evaluation}

\begin{figure}[thbp!] 
\centering
\includegraphics[width=\linewidth]{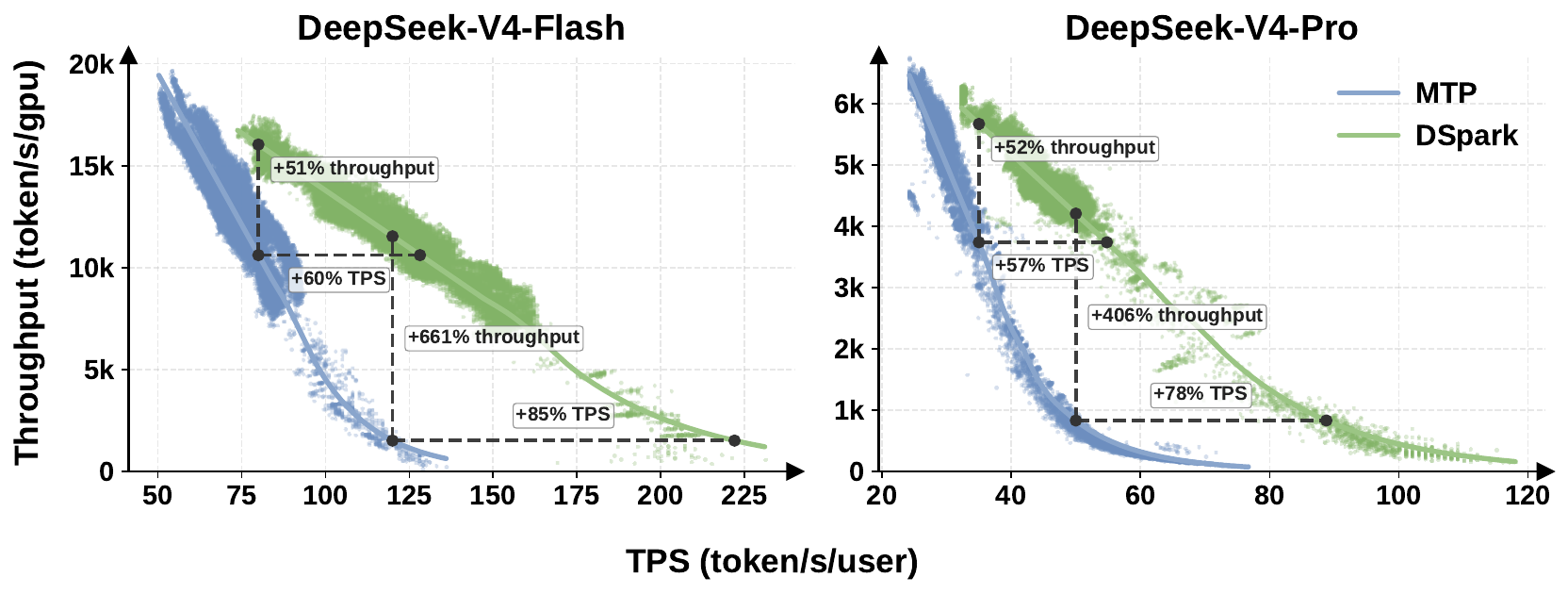} 
\caption{
\textbf{Throughput vs.\ TPS.} 
Aggregate output token throughput against per-request generation speed (tok/s/user) under live traffic. 
In our production deployment, DSpark improves the observed throughput–interactivity frontier relative to the MTP-1 baseline under the measured traffic and engine configurations.
} 
\label{fig:prod-interactivity} 
\end{figure} 

We evaluate DSpark-5 (configured with a maximum draft length of $\gamma=5$) against the MTP-1~\citep{liu2024deepseek} baseline within the production serving engines of DeepSeek-V4-Flash~(preview) and DeepSeek-V4-Pro~(preview).
MTP-1 represents the former production setup, having been superseded by DSpark two weeks following the DeepSeek-V4-preview release.
This single-token setup was historically maintained in production because deploying a static multi-token drafter (e.g., MTP-3/5) strictly degrades aggregate throughput under high concurrency due to excessive verification overhead. 
Therefore, comparing DSpark against this established baseline directly demonstrates its ability to safely unlock the performance potential of larger draft blocks in dynamic serving environments. 
In all figures, the scatter points represent raw telemetry data sampled directly from live user traffic, capturing complex, real-world request distributions, while the solid lines represent the fitted performance frontiers.

\paragraph{The Serving Pareto Frontier.}
\autoref{fig:prod-interactivity} illustrates the trade-off between aggregate system throughput and per-user generation speed (interactivity).
To quantify DSpark's behavior under practical deployment constraints, we evaluate the system at several interactivity SLA anchors. Here, an SLA (Service Level Agreement) specifies the minimum per-user generation speed (in tokens per second) that the system must guarantee.

For the V4-Flash engine, we evaluate the system at SLA anchors of 80 and 120 tok/s/user. 
At the moderate 80 tok/s/user SLA, DSpark improves aggregate throughput by 51\% over the MTP-1 baseline.
The stricter 120 tok/s/user SLA represents a qualitatively different regime: under this constraint, the single-token MTP-1 baseline approaches its operational boundary and can sustain only a very small concurrent batch.
Consequently, the relative throughput ratio at this point is numerically large, with DSpark achieving a nominal 661\% higher aggregate throughput.
We therefore interpret this high-SLA point primarily as evidence that DSpark extends the feasible interactivity frontier, rather than as a representative multiplicative speedup over a well-utilized baseline.
At matched practical throughput levels, which provide a more stable comparison, DSpark accelerates per-user generation speeds by 60\% to 85\%.

The V4-Pro deployment shows the same pattern.
At the moderate 35 tok/s/user SLA, DSpark improves aggregate throughput by 52\%.
At the stricter 50 tok/s/user SLA, MTP-1 again enters a low-concurrency regime, yielding a nominal 406\% relative throughput advantage for DSpark.
As with V4-Flash, we treat this point as an indication that DSpark sustains useful throughput under an interactivity target that the baseline cannot efficiently support.
At matched system capacities, DSpark delivers 57\% to 78\% faster per-user generation.
Overall, these results show that DSpark shifts the observed throughput--interactivity frontier outward: it improves throughput in moderate-SLA regimes and, more importantly, preserves non-degenerate serving capacity under strict interactivity constraints.

\paragraph{Throughput Dynamics under Load.}

\begin{figure}[thbp!]   
\centering  
\includegraphics[width=0.9\linewidth]{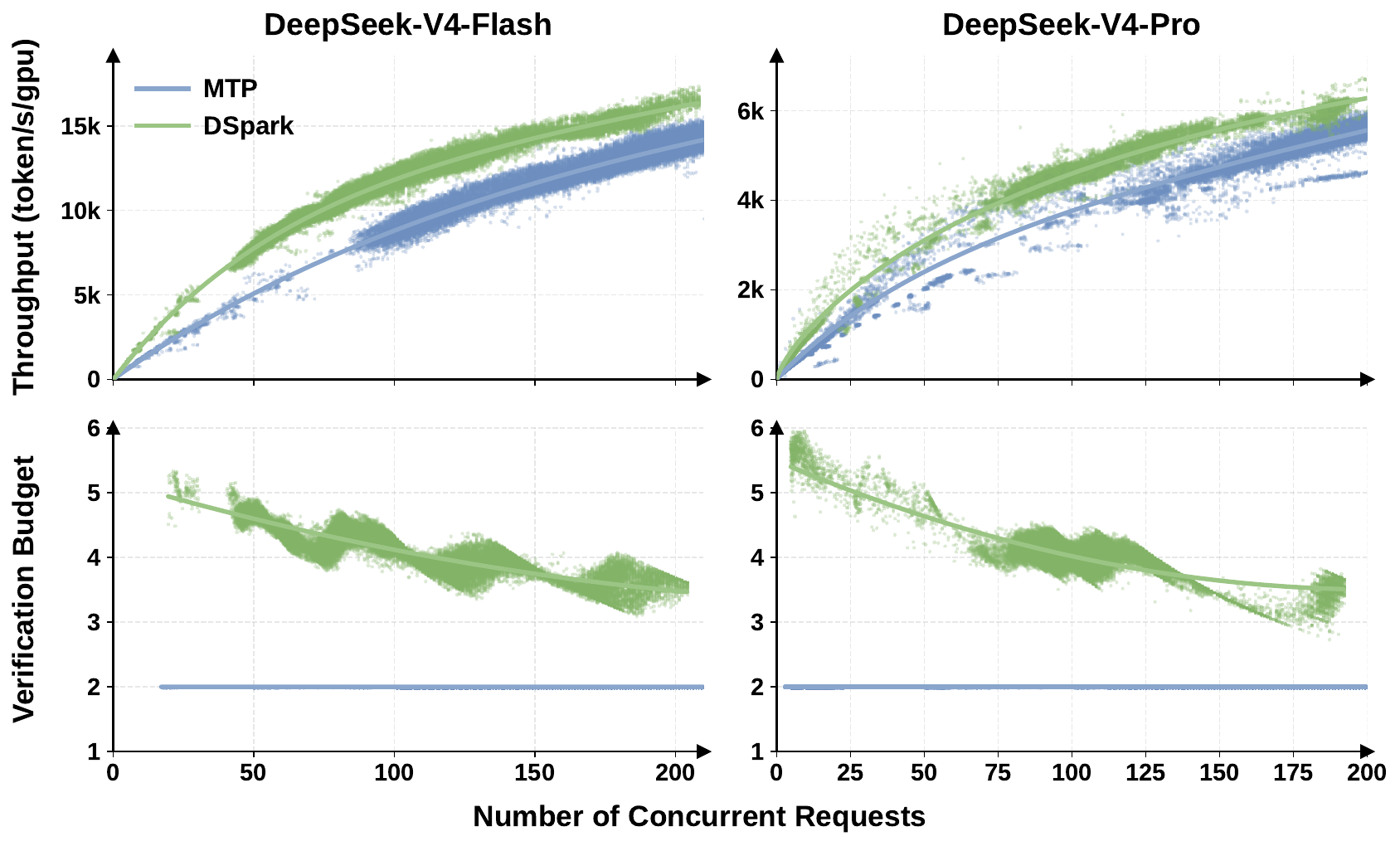}   
\caption{\textbf{Load-adaptive throughput and verification budgets.} Top row (a, b): Aggregate output throughput across varying levels of system concurrency. Bottom row (c, d): The average target verification budget allocated per request. As concurrent load increases, the dynamic scheduler automatically restricts the per-request verification length to prevent resource contention.}   
\label{fig:prod-scalability}   
\end{figure}

\autoref{fig:prod-scalability} analyzes the underlying mechanism driving these gains by plotting aggregate throughput (top row) and the dynamic verification budget (bottom row) against system concurrency.
\begin{itemize}
    \item Under the moderate concurrency regimes typical of our production deployment (fewer than 200 concurrent requests for V4-Flash and 150 for V4-Pro), the hardware-aware scheduler leverages available target compute capacity by allocating longer verification budgets, expanding from MTP-1's static 2 tokens to roughly 4--6 tokens per request. This extended verification yields more accepted tokens per forward pass, directly contributing to the throughput gains observed on the Pareto frontier.
    \item As system concurrency scales and target capacity saturates, the scheduler dynamically restricts this budget. The average verification length decreases smoothly with load, ensuring that low-confidence draft tokens are pruned before they consume critical batch capacity. This load-aware behavior stabilizes production deployment: DSpark maximizes the utility of idle compute under light traffic, while effectively preserving critical batch capacity under heavy traffic.
\end{itemize}

\paragraph{Limitations.}Although the prefix scheduler minimizes wasted target-model verification, DSpark still incurs a fixed draft-side cost to generate the initial $\gamma$-token block via the parallel backbone. For complex queries with inherently low acceptance rates, this upfront drafting compute is unrecoverable. Future optimizations could introduce difficulty-aware early exiting within the draft model, enabling such requests to bypass full-block generation.
\section{Related Work}
\label{sec:related_work}

\paragraph{Speculative Decoding Algorithms.}
Speculative decoding accelerates autoregressive generation by decoupling token proposal from verification. Building on early blockwise methods~\citep{stern2018blockwiseparalleldecodingdeep,sun-etal-2021-instantaneous,ge2022lossless,xia2023speculativedecodingexploitingspeculative}, modern approaches employ rejection sampling to exactly preserve the target model's distribution~\citep{chen2023accelerating,pmlr-v202-leviathan23a}. Because inference speedup directly depends on the drafter's efficiency and accuracy, extensive research has focused on optimizing its architecture. Beyond using standalone small language models~\citep{chen2023accelerating,pmlr-v202-leviathan23a}, subsequent work integrates multi-token heads or feature extrapolators directly into the target model~\citep{pmlr-v235-cai24b,ankner2024hydrasequentiallydependentdraftheads,pmlr-v235-li24bt,li-etal-2024-eagle,li2026eagle,zhang2025learningharmonizedrepresentationsspeculative,pmlr-v235-gloeckle24a,liu2024deepseek,cai2025fastmtpacceleratingllminference,eldenk2026attentiondriftautoregressivespeculative}. Other strategies include self-speculation via early exits~\citep{Zhang_2024,Elhoushi_2024,NEURIPS2024_16336d94,xia2025swiftontheflyselfspeculativedecoding}, dynamic vocabulary compression~\citep{zhao2025fr,williams2026speculative}, prompt lookup~\citep{saxena2023prompt,somasundaram2024pldacceleratingllminference}, suffix automata~\citep{hu2024samdecodingspeculativedecoding}, and retrieval~\citep{he2023rest,shen2026draftlessretrievemore}.
To remove the sequential bottleneck of the drafting itself, a line of research proposes parallel or blockwise generation, including Medusa~\citep{pmlr-v235-cai24b}, P-EAGLE~\citep{hui2026peagleparalleldraftingeaglescalable}, PARD~\citep{an2026pard,an2026pard2}, DART~\citep{liu2026dartdiffusioninspiredspeculativedecoding} and DFlash~\citep{chen2026dflash}.
DDTree, TAPS and JetSpec then extend the draft chain to verifiable trees~\citep{ringel2026ddtree,wang2026tapstargetawareprefixtree,hu2026jetspecbreakingscalingceiling}. 
Concurrent efforts include: Domino~\citep{huang2026dominodecouplingcausalmodeling} introduces a CausalEncoder conceptually similar to our RNN Head; DFlare~\citep{zhang2026dflarescalingdraftcapacity} addresses conditioning bottlenecks via layer-wise fusion.

\paragraph{System-Aware Scheduling for Speculative Decoding.}
Beyond drafter architecture, another line of work focuses on determining the optimal number of speculative tokens to generate or verify in each round. To this end, various approaches adapt draft lengths on the fly using confidence heuristics~\citep{mamou2024dynamicspeculationlookaheadaccelerates,li-etal-2024-eagle, du2024glide,liu2026talon,wen2026specboundadaptiveboundedselfspeculation}, learned acceptance predictors~\citep{huang2024specdec++, zacks9172026automtpvllm}, or bandit-style policies~\citep{liu2026notabanditprovablynoregretdrafter}. Furthermore, recognizing speculative decoding as inherently a system-level scheduling problem, recent works optimize overall goodput and latency by adjusting speculation budgets according to real-time system load and request priority~\citep{angelslim2026dcut, liu2024turbospec,10.1145/3772052.3772239,li2026nightjardynamicadaptivespeculative,wu-etal-2025-tetris,hu2026echo,miao2024specinfer,sadhukhan2025magicdecbreakinglatencythroughputtradeoff}.

\paragraph{Parallel Generation.}
Models that generate tokens in parallel offer a decoding latency nearly independent of output length, making them an attractive alternative to autoregressive decoding.
Non-Autoregressive Transformers \citep[NATs,][]{gu2018nonautoregressive} pioneered this direction by predicting all positions independently in a single pass.
However, this forces the model to average over all plausible modes, often producing outputs that mix fragments from different valid sequences. 
Two broad lines of work have emerged to address this limitation.
One direction retains the single-pass architecture but changes what the model sees or how it is trained: introducing latent variables as conditioning input to steer all positions toward a consistent output~\citep{gu2018nonautoregressive,pmlr-v80-kaiser18a,ma-etal-2019-flowseq}, or relaxing the training objective so that the model focuses on producing a single coherent output rather than modeling the full distribution over all valid alternatives~\citep{shao-etal-2021-sequence,pmlr-v139-du21c,qian-etal-2021-glancing,shao2023beyond}.
The other direction reintroduces limited sequential dependency through iterative re-prediction~\citep{ghazvininejad-etal-2019-mask,austin2021structured,li2022diffusionlm}, block-level autoregression~\citep{wang-etal-2018-semi-autoregressive,arriola2025block}, or structured output layers such as CRF~\citep{NEURIPS2019_74563ba2}, CTC~\citep{libovicky-helcl-2018-end,saharia-etal-2020-non}, HMM~\citep{pmlr-v162-huang22m}, and PCFG~\citep{gui2023nonautoregressive}.

Speculative decoding places a further demand that the drafter must provide exact per-token probabilities for the rejection sampling rule. 
Most techniques above cannot readily provide such probabilities due to iterative refinement, latent marginalization, or global normalization.
For instance, in a design closely related to ours, CRF-NAT~\citep{NEURIPS2019_74563ba2} also places a sequential module over parallel hidden states, but its globally normalized partition function prevents exact per-token probability computation. Similarly, when adapting the CTC output layer to parallel speculative decoding, CTC-drafter~\citep{NEURIPS2024_a79054a9} is restricted to greedy verification due to the latent marginalization of alignment paths. DSpark circumvents these limitations by keeping the sequential correction local, so per-token probabilities remain exact softmax evaluations.

\section{Conclusion}
\label{sec:conclusion}

In this paper, we present DSpark, a speculative decoding framework designed to overcome the structural and system-level bottlenecks of large language model inference in high-concurrency production environments. Algorithmically, DSpark introduces a semi-autoregressive generation paradigm---coupling a computationally heavy parallel backbone with a lightweight sequential head---to mitigate the rapid suffix decay of independent parallel drafters. At the system level, we formulate verification length selection as a global throughput maximization problem, employing a hardware-aware prefix scheduler that dynamically tailors the target model's verification budget based on calibrated survival probabilities and real-time engine load. Extensive offline evaluations demonstrate that DSpark substantially outperforms state-of-the-art autoregressive and parallel baselines across diverse domains. Furthermore, its real-world deployment within the DeepSeek-V4 validates its practical value in production serving: by intelligently managing verification overhead, DSpark sustains robust concurrency under heavy load, consistently accelerates per-user generation speeds, and shifts the Pareto frontier of LLM serving outward.
\bibliography{ref}

\newpage
\appendix
\section*{Appendices}
\section{Counterexample: Selection Bias Without Early-Stopping}
\label{app:counterexample}
We provide a simple counterexample to illustrate how an offline global search, i.e., operating without the \texttt{break} condition in \autoref{alg:prefix-scheduler}, violates the non-anticipating property required by lossless speculative decoding. Formally, the admission event for the $k$-th draft token, $\ell_r \ge k$, must be determined by scheduler-visible information available before the token $x_{r,k}$ is sampled. It must not depend on the realization of $x_{r,k}$ itself.
Consider a scenario with a single request ($R=1$) and maximum draft length ($\gamma=2$). Suppose the pre-token confidence for the first position is $a_1 = 0.8$, and the profiled capacity curve is
\[
\mathrm{SPS}(1) = 1.0,\qquad
\mathrm{SPS}(2) = 0.5,\qquad
\mathrm{SPS}(3) = 0.45.
\]
The expected throughputs for verifying $0$ and $1$ draft tokens are
\begin{align*}
\Theta_0 &= 1 \cdot \mathrm{SPS}(1) = 1.0, \\
\Theta_1 &= (1 + 0.8) \cdot \mathrm{SPS}(2) = 0.9.
\end{align*}
Without early-stopping, the scheduler proceeds to evaluate $\Theta_2$ before committing any admission decisions. Because the Markov confidence head uses the previously sampled token, the next confidence score $c_2$ explicitly depends on the realization of $x_1$. Consequently, the second-prefix survival probability
\[
a_2 = a_1 c_2
\]
also depends on $x_1$. Consider two possible realizations of $x_1$:
\begin{itemize}
\item \textbf{Case 1 ($x_1$ yields a high $c_2$):}
Suppose $x_1$ results in $c_2 = 0.9$. Then
\[
a_2 = 0.8 \times 0.9 = 0.72.
\]
The expected throughput for length $2$ is
\[
\Theta_2 = (1 + 0.8 + 0.72) \times 0.45 = 1.134.
\]
Since $\Theta_2$ is the global maximum among $\{1.0, 0.9, 1.134\}$, the scheduler returns $\ell=2$. The first token $x_1$ is admitted into the verification prefix.
\item \textbf{Case 2 ($x_1$ yields a low $c_2$):}
Suppose $x_1$ results in $c_2 = 0$. Then
\[
    a_2 = 0.
\]
The expected throughput for length $2$ is
\[
    \Theta_2 = (1 + 0.8 + 0) \times 0.45 = 0.81.
\]
Here, the global maximum remains $\Theta_0 = 1.0$, so the scheduler returns $\ell=0$. The first token $x_1$ is not admitted into the verification prefix.
\end{itemize}
Thus, the admission of the first draft token dynamically depends on the value of the first draft token itself. This retrospective dependence introduces selection bias: the scheduler favors tokens that lead to highly confident continuations, even though the admission decision for $x_1$ should have been made before observing $x_1$.
We now make the distributional bias explicit. Let the vocabulary be $\{A,B\}$, and consider the target and draft distributions at the first position:
\[
p_{\mathrm{t}}(A)=0.7,\qquad p_{\mathrm{t}}(B)=0.3,
\]
\[
p_{\mathrm{d}}(A)=0.5,\qquad p_{\mathrm{d}}(B)=0.5.
\]
The standard speculative acceptance probability at the first position is
\[
\sum_{x\in\{A,B\}} \min\bigl(p_{\mathrm{t}}(x),p_{\mathrm{d}}(x)\bigr)
=
\min(0.7,0.5)+\min(0.3,0.5)
=
0.8,
\]
matching the assumed value ($a_1=0.8$).
Suppose the retrospective scheduler behaves as above: $x_1=A$ yields a high continuation confidence and hence $\ell=2$, while $x_1=B$ yields a low continuation confidence and hence $\ell=0$. Then the first output token is distributed as follows. If $x_1=A$, the draft token is admitted and accepted with probability
\[
\min\left(1,\frac{p_{\mathrm{t}}(A)}{p_{\mathrm{d}}(A)}\right)
=
\min\left(1,\frac{0.7}{0.5}\right)
=
1,
\]
so the output token is $A$. If $x_1=B$, the draft token is not admitted; the target model instead generates a fresh token from $p_{\mathrm{t}}$. Therefore,
\[
\Pr(Y=A)
=
\Pr(x_1=A)\cdot 1
+
\Pr(x_1=B)\cdot p_{\mathrm{t}}(A)
=
0.5 + 0.5 \times 0.7
=
0.85,
\]
and hence
\[
\Pr(Y=B)=0.15.
\]
This output distribution ($(0.85,0.15)$) differs from the target distribution ($(0.7,0.3)$), proving that the retrospective scheduler is not lossless.
The early-stopping mechanism prevents this issue in the causal greedy scheduler. Since $\Theta_1 < \Theta_0$, the scheduler halts immediately and returns $\ell=0$ before evaluating any continuation-dependent quantity such as $c_2$. The admission decision for the first position therefore depends only on pre-token information and cannot be biased by the realization of $x_1$. This restores the non-anticipating property required by the standard losslessness argument.

\end{CJK*}
\end{document}